\documentclass{article} 
\usepackage[preprint]{neurips_2026}
\setcitestyle{numbers,square,comma}
\usepackage[utf8]{inputenc} 
\usepackage[T1]{fontenc}

\usepackage{url} 
\usepackage{booktabs} 
\usepackage{amsfonts} 
\usepackage{nicefrac} 
\usepackage{microtype} 
\usepackage[table]{xcolor}   
\usepackage{enumitem}
\usepackage{subcaption}
\usepackage{graphicx} 
\usepackage{caption} 
\usepackage{multirow} 
\usepackage{float} 
\usepackage{placeins} 
\usepackage[accsupp]{axessibility}
\usepackage[colorlinks,urlcolor=blue,linkcolor=blue,citecolor=blue]{hyperref}

 \title{Point as Skeleton: Accumulated Point Cloud Enhanced Autoregressive Generation for Closed-Loop Autonomous Driving Simulation}

\usepackage{pifont}
\newcommand{\mailmark}{\raisebox{-0.02em}{\scalebox{0.82}{\ding{41}}}}
\newcommand{\corrmark}{\textsuperscript{\mailmark}}

\author{%
\normalfont
\fontsize{9.6}{11.8}\selectfont
\begin{tabular}{@{}c@{}}
Songbur Wong$^{1}$,
Xiaosong Jia$^{3}$\corrmark,
Junqi You$^{1}$,
Bo Zhang$^{4}$,
Pei Xu$^{2}$ \\[2pt]
Renqiu Xia$^{1}$,
Yuping Qiu$^{2}$,
Shaofeng Zhang$^{5}$,
Zelin Zhao$^{6}$,
Xuechao Yan$^{2}$ \\[2pt]
Yuchen Zhou$^{3}$,
Yurui Chen$^{2}$,
Wen Guo$^{2}$,
Hang Xu$^{2}$,
Junchi Yan$^{1}$\corrmark \\[5pt]
$^{1}$Shanghai Jiao Tong University \quad
$^{2}$Yinwang Intelligent Technology Co., Ltd. \\
$^{3}$Fudan University \quad
$^{4}$Shanghai Artificial Intelligence Laboratory \\[2pt]
$^{5}$University of Science and Technology of China \quad
$^{6}$Georgia Institute of Technology \\[2pt]
\mailmark~Corresponding authors
\end{tabular}
\vspace{-5mm}
}

\begin{document}

\maketitle

\begin{abstract}
Evaluating end-to-end autonomous driving (E2E-AD) remains challenging, as existing driving simulation methods often trade off closed-loop interactivity (e.g., CARLA) and real-world visual fidelity (e.g., nuScenes).
We present \textbf{\emph{Point as Skeleton}}, a generative sensor simulation framework for state-updated autoregressive driving video generation, in which an autoregressive generator synthesizes visual observations from step-wise updated ego states, actor states, scene maps, and point-cloud skeleton conditions.
To support closed-loop rollout, we introduce Reset-and-Roll, which adapts rolling diffusion inference to simulation by preventing future-conditioned latent states from being committed across simulation steps.
To stabilize error accumulation during step-wise autoregressive rollout, we introduce point-cloud skeletons that decouple foreground and background assets and project them into camera-view painted-point and template-depth conditions, providing appearance and geometric cues.
We further implement a nuPlan-based renderer-level closed-loop generative interface for evaluating generation under ego deviations from the original log.
Experiments on nuScenes and nuPlan show that \textit{Point as Skeleton} improves autoregressive generation quality during closed-loop rollout, demonstrating its potential for visually faithful closed-loop driving simulation. The code is available at \url{https://github.com/krauwu/point-as-skeleton}.
\end{abstract}
\vspace{-3mm}

\section{Introduction}

End-to-end autonomous driving (E2E-AD)~\cite{uniad,li2024think2drive,yang2025drivemoemixtureofexpertsvisionlanguageactionmodel,jia2025drivetransformer,jia2023driveadapter} has attracted increasing attention in recent years. A central challenge in evaluating E2E-AD systems is to provide sensor observations that respond to the actions selected by the driving policy. 
Open-loop benchmarks such as nuScenes~\cite{nuscenes} provide realistic sensor data, but the observations are fixed by the recorded trajectory and cannot react to policy-induced actions. In contrast, closed-loop simulators such as CARLA~\cite{carla,zhang2023resimad,jia2024bench} provide interactive feedback, but their rendered sensor appearance can still differ from real-world camera data. 
This motivates generative sensor simulation, where a learned generative model synthesizes camera observations from simulator-provided ego states, actor states, and scene conditions.

However, closed-loop simulation goes beyond standard driving-scene reconstruction~\cite{streetgs} and generation~\cite{drivedreamer,genad}. The ego vehicle may deviate from the logged trajectory, while the planner and traffic simulator update online.
The generator therefore needs to render multi-view observations under off-log motions and repeatedly refreshed conditions. This introduces two additional requirements:

\begin{figure}[t!]
\includegraphics[width=\textwidth]{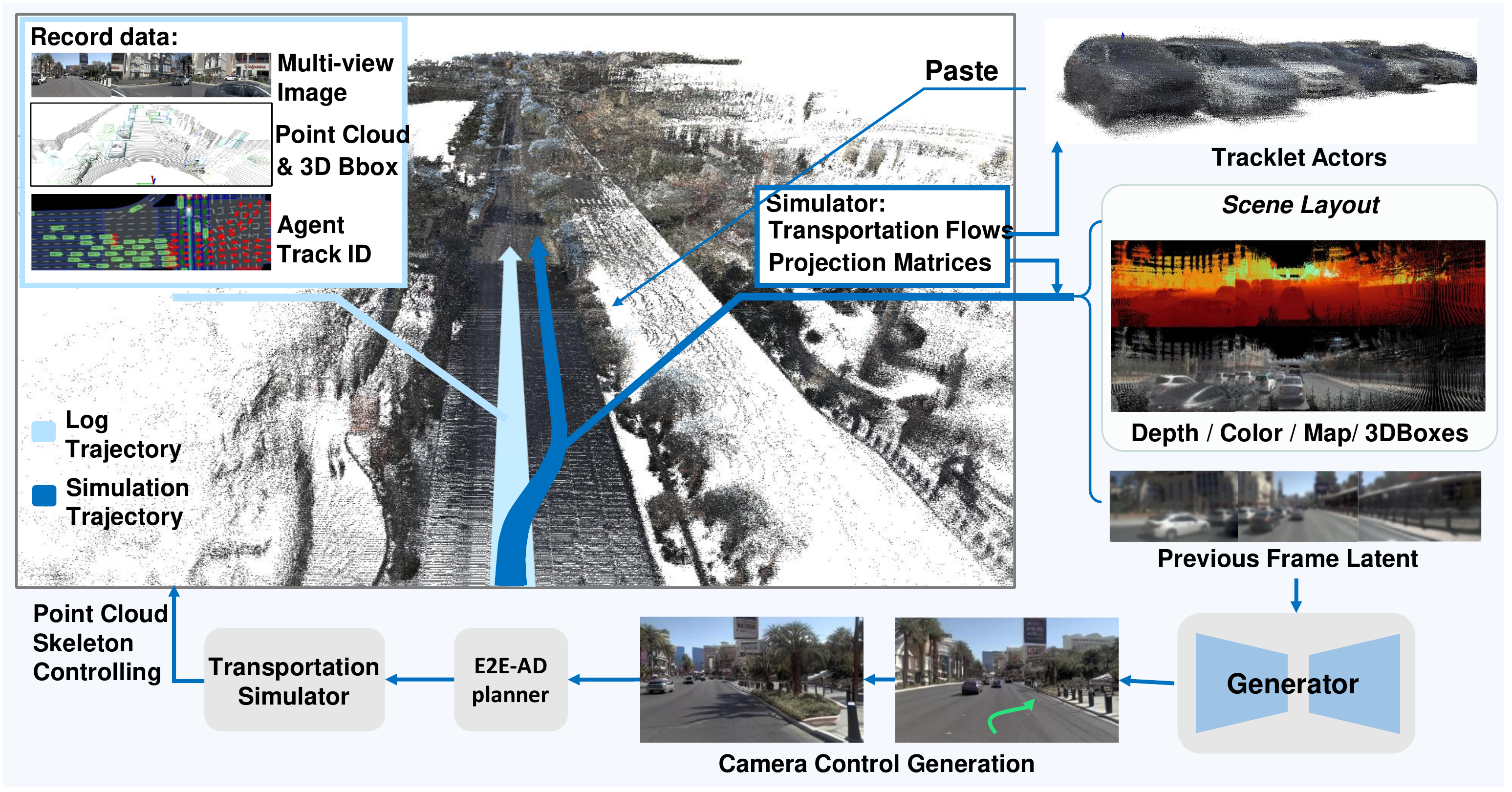}
  \caption{
  \textbf{\textit{Point as Skeleton} Generative Simulator.}
  We utilize offline logs to build a foreground-background decoupled Point Cloud Skeleton.
  At each time step, the evaluated E2E-AD model predicts an action, and the traffic simulator updates the ego state and surrounding actor states accordingly.
  Based on the updated actor information, the Point Cloud Skeleton is edited and projected back to the camera views as multiple conditions.
  Finally, the generator synthesizes the latest sensor observations, which are fed back to the E2E-AD model for the next-step decision.
  }
  \vspace{-6mm}
  \label{fig:teaser}
\end{figure}

\noindent\textbf{(I) Spatiotemporal Extrapolation.}
Closed-loop policies may deviate from the logged motion in speed or heading, exposing vehicles and pedestrians that are not observed in the recorded images. This is challenging for classical reconstruction pipelines~\cite{omnire}. Recent methods~\cite{recamdriving,r3d2} use generative models as reconstruction post-processing to improve visual plausibility, but they require scene- or frame-specific processing, reducing their ease of deployment.

\noindent\textbf{(II) Instantaneous Interactivity.}
Closed-loop simulation updates the planner action at every step, requiring the simulator to synthesize the next observation from the latest scene state.
Reconstruction-based renderers can be repeatedly queried once a scene is built.
However, reconstruction-generation hybrids~\cite{freevs,mirage} and urban-scene generators~\cite{drivingsphere,dist4d} usually assume pre-defined conditions over a full clip. This calls for \textbf{frame-wise autoregressive generation} with step-wise condition updates.

Existing studies~\cite{you2024bench2drive,diffusionforcing} show that directly applying full-clip video generators to frame-wise autoregressive rollout causes severe error accumulation (Fig.~\ref{fig:observation}). 
We therefore propose \textit{Point as Skeleton}, which uses point-cloud assets accumulated once offline as stabilizing conditions for autoregressive generation.
The resulting simulator uses an autoregressive video generator as the visual backbone, while the point skeleton provides editable geometric cues to reduce error accumulation.
To realize this framework, we make the following designs.

\noindent\textbf{Reset-and-Roll autoregressive generation.}
We observe that Rolling Diffusion~\cite{rollingdiffusionmodel} inference is better suited to highly dynamic driving scenes, but its input-output timestamps are not aligned with the simulator interface.
A simulator step expects the generator to consume the current scene condition and immediately return the next observation, whereas rolling diffusion emits the earliest observation in its rolling window after using lookahead conditions.
Reset-and-Roll bridges this gap by allowing rolling-style denoising while resetting the committed latent state before future-conditioned information is passed to the next simulation step.

\noindent\textbf{Point-skeleton conditions.}
We design two point-cloud conditions to stabilize autoregressive generation.
The painted point cloud~\cite{vora2020pointpainting} provides pixel-level appearance guidance and suppresses visual degradation during long rollout.
However, painted points may suffer from reconstruction inconsistency when actor instances are only partially scanned, motivating a parallel depth condition as a stable geometric anchor.
The template-based depth branch replaces foreground actors with category-level point templates, providing stable depth guidance for autoregressive rollout.

\noindent\textbf{nuPlan SimGen.}
We implement a closed-loop generative interface on nuPlan~\cite{nuplan}.
The interface attaches our visual generator to the nuPlan traffic simulation core, allowing the ego vehicle to change speed and heading according to user-defined commands or nuPlan planners.
It enables evaluation of autoregressive generation under trajectories that deviate from the original log and provides projected point-based semantic labels for perception-oriented evaluation.

\section{Related Works}

\subsection{Driving World Models}

Conditional diffusion models have shown strong controllability in visual generation~\cite{controlnet,ominicontrol2}.
As conditioning signals extend from text and semantic maps to 3D boxes, HD maps, ego motion, and camera geometry, these models have become increasingly applicable to driving-scene generation~\cite{bevcontrol,magicdrive,magicdrivedit,panacea}.
A line of work studies driving video generation as predictive world modeling~\cite{genad,vista,drivedreamer,drivedreamer2,drivingsphere,unimlvg,dist4d}.
These methods synthesize future scene observations under given actions or scene-level conditions, and introduce driving-specific designs such as structured trajectory tokens~\cite{genad}, action-controllable future prediction~\cite{vista}, and depth or occupancy conditions~\cite{dist4d,drivingsphere}.
Their main focus is to generate plausible future clips under a specified condition sequence.

More recent works further explore autoregressive driving world models.
Epona~\cite{epona} formulates driving world modeling as next-token prediction and jointly studies generation and planning.
OmniNWM~\cite{omninwm} extends this direction toward panoramic navigation world modeling with state, action, and reward awareness.
These methods indicate a shift from fixed-length video synthesis toward longer-horizon and more interactive world modeling.
However, their design objective still differs from closed-loop simulation.
They are generally designed to predict future pixels and actor motions from historical observations and action-related inputs, rather than to render observations under externally updated scene layouts.
As a result, fine-grained control over individual participants is less direct than in explicit layout-conditioned simulation, where 3D boxes, HD maps, and simulator states can be updated at every step.

\subsection{Driving Scene Simulation}

Driving scene simulation aims to provide sensor observations for interactive evaluation of autonomous-driving policies.
Classical simulators such as CARLA~\cite{carla} support closed-loop interaction, but their rendered appearance often differs from real-world camera data.
Reconstruction-based methods improve visual fidelity by reconstructing urban scenes from recorded logs~\cite{streetgs,emernerf}, yet they are most reliable near the logged trajectory.
When the ego vehicle deviates from the recorded motion, unseen regions and weakly observed actors become difficult to render faithfully.

Recent methods introduce generative priors to alleviate this limitation.
FreeVS~\cite{freevs} improves free-viewpoint synthesis under shifted driving trajectories, ReCamDriving~\cite{recamdriving} conditions generation on reconstructed scenes, and R3D2~\cite{r3d2} refines projected dynamic instances with generative models.
These reconstruction-generation hybrids improve visual plausibility beyond logged views, but they still rely on scene-specific reconstruction, object-level editing, or post-processing modules.
Therefore, extending them to continuous closed-loop rollout with online condition updates is not straightforward.

Generative simulation systems move one step further by using learned world models as simulator engines.
DriveArena~\cite{drivearena} builds a closed-loop generative simulation platform, where generated observations are consumed by a driving agent, the resulting trajectory updates the scene layout, and the updated layout is fed back to the generator.
Closed-source systems such as Waymo World Model~\cite{waymoworldmodel} and GAIA-3~\cite{gaia3} further demonstrate controllable autonomy evaluation with action-, layout-, and scenario-conditioned generation.
These systems suggest that driving simulation is moving from graphics- or reconstruction-centered pipelines toward generation-centered closed-loop simulators.
Nevertheless, open autoregressive video-generation designs that can be readily integrated with closed-loop driving systems remain limited.

\section{Method}

\textbf{Overview.}
We build a frame-wise generative rendering interface for closed-loop-compatible autonomous driving simulation: each updated simulator state defines the next scene layout, and the generator synthesizes the corresponding camera observation.
Our method includes (i) an autoregressive video diffusion model with diffusion-forcing training and \emph{Reset-and-Roll inference} for stable rollouts (Sec.~\ref{framework}), (ii) a \emph{Point Cloud Skeleton} projected as color and depth conditions for view-consistent geometry (Sec.~\ref{data}), and (iii) nuPlan-SimGen, an interactive generative rendering interface and evaluation
protocol for ego-deviated trajectories (Sec.~\ref{sim_geneval}).

\subsection{Autoregressive Framework}
\label{framework}

\begin{figure}[t!]
\includegraphics[width=\textwidth]{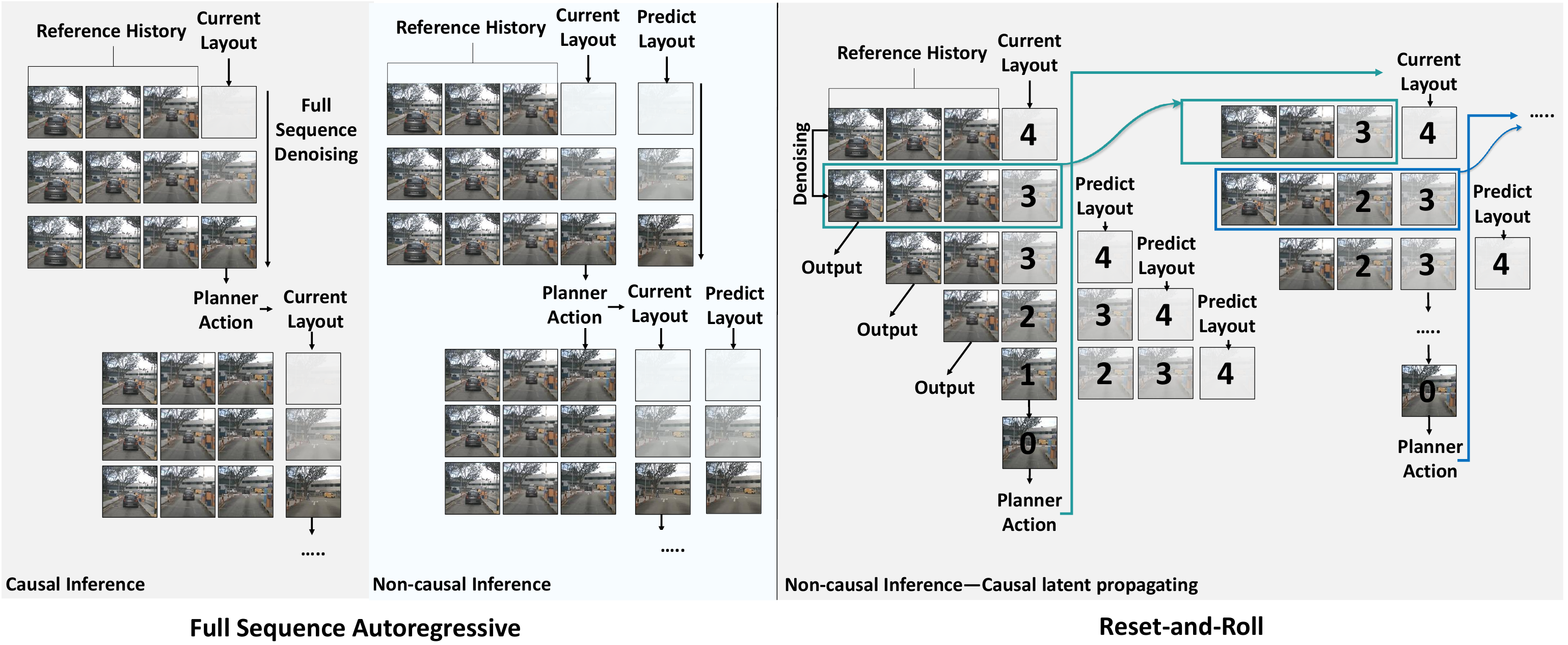}
  \caption{
  \textbf{Reset-and-Roll Framework.} Unlike full-sequence denoising~\cite{drivearena,you2024bench2drive,epona}, Reset-and-Roll supports past-to-future conditioned denoising, while preventing the non-causal guidance from accumulating errors across simulation steps. In our implementation, a world prediction model~\cite{unimlvg} trained with standard diffusion forcing~\cite{dfot} is sufficient for Reset-and-Roll
inference.
  }
  \label{fig:genframe}
  \vspace{-3mm}
\end{figure}

In closed-loop simulation, the generator acts as a step-wise visual renderer.
At simulation step \(t\), the planner outputs a short-horizon trajectory as
discrete ego waypoints. The simulator commits only the next executable waypoint
as the current ego state, while the remaining waypoints are treated as
short-horizon forecasts. We therefore distinguish between \emph{committed
states} and \emph{forecast states}. Committed states include the generated
history, the current ego state, the current actor states, and the simulator
layout \(\mathcal{L}^{\mathrm{sim}}_t\) at step \(t\). Forecast states include
future ego waypoints and their corresponding future layouts
\(\hat{\mathcal{L}}_{t+1:t+K}\), which may be revised after the policy receives
the newly generated observation.

As shown in the left part of Fig.~\ref{fig:teaser}, the traffic simulator places
the committed ego and actor states on the map to construct the current layout
\(\mathcal{L}^{\mathrm{sim}}_t\). Future layouts are constructed in the same
format, but they remain mutable forecasts rather than simulator states to be
inherited. The video generator is therefore invoked at \textbf{every simulation
step}, conditioned on the generated history, the current committed layout, and
the forecast layouts within the generation window.

\begin{figure}[t!]
\includegraphics[width=\textwidth]{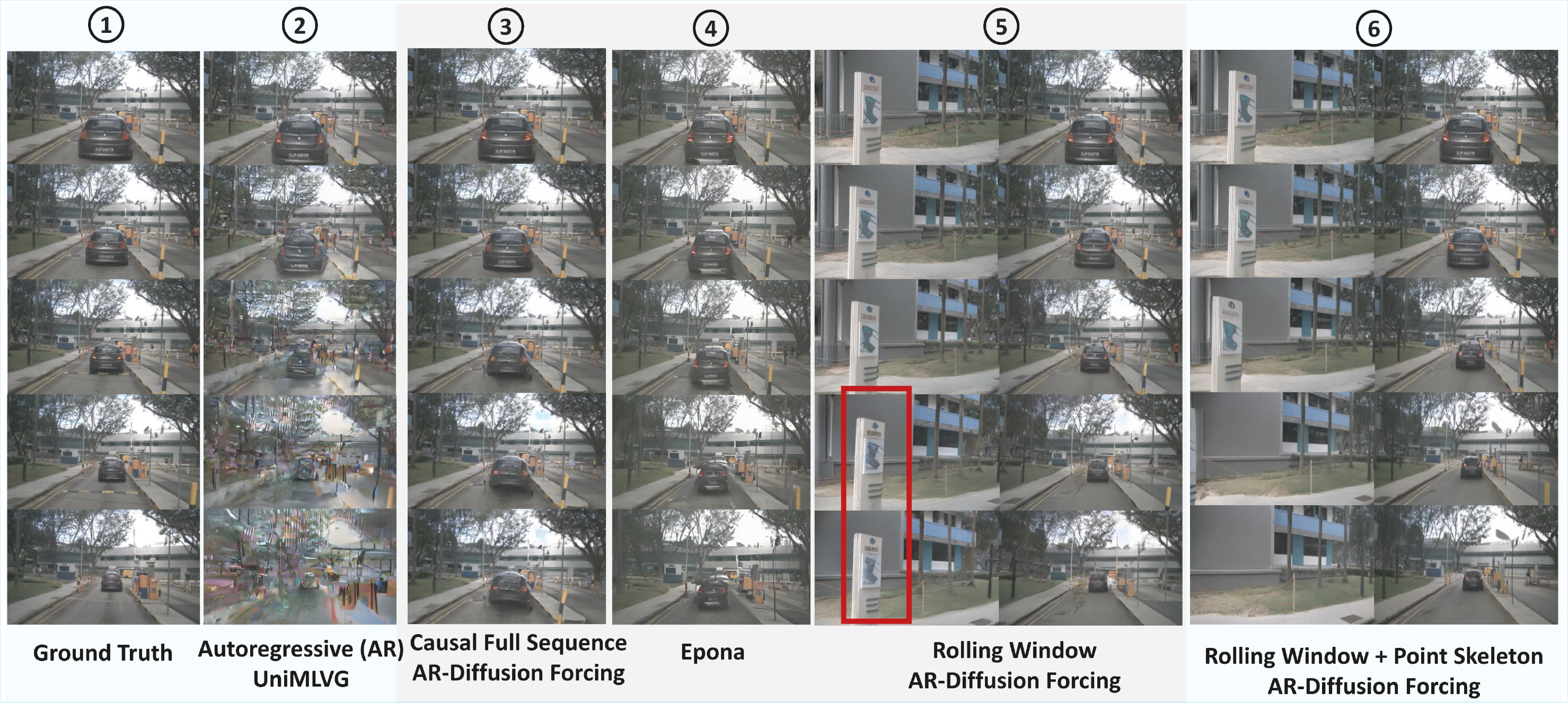}
  \caption{
  \textbf{Qualitative Observation.} Results are generated frame-wise autoregressively on the first nuScenes-val sequence. Note that sequence (3) and (5) share the same checkpoint from diffusion forcing training, and sequence (2) and (4) used official model. Both sequences (3) and (4) use full-sequence causal AR and show vehicle stagnation. Sequence (5) shows static background moves with the ego vehicle, suggesting weak geometric consistency in frame-wise autoregressive generation.
  }
  \label{fig:observation}
    \vspace{-4mm}
\end{figure}

This setting casts generative simulation as a frame-wise autoregressive video generation problem: each synthesized observation is returned to the policy and affects the next simulator query. Without dedicated training or rollout design, autoregressive diffusion suffer from severe error accumulation~\cite{you2024bench2drive,diffusionforcing}, making standard autonomous-driving video generators difficult to use as interactive renderers, as shown in Fig.~\ref{fig:observation}-(2). Existing works alleviate this issue by noising frames during training and inference~\cite{you2024bench2drive,omninwm}, or by simulating inference-time histories during training~\cite{epona}. However, when deployed as closed-loop renderers, full-sequence inference still suffers from two failure modes.

\textbf{Causal full-sequence.}
As shown in Fig.~\ref{fig:genframe}, if the generator avoids future layouts, the next-frame distribution is
marginalized over possible future motions:
\[
p(x_{t+1}\mid h_t)=
\int p(x_{t+1}\mid h_t,f_t)p(f_t\mid h_t)\,df_t ,
\]
where \(h_t\) denotes the committed history and current simulator layout, and
\(f_t\) denotes future layout trajectories. This removes explicit motion cues
during denoising; thus low-speed or occluded actors are often generated
conservatively, causing weak or static motion in rollout
(Fig.~\ref{fig:observation}-(3),(4)).

\textbf{Non-causal full-sequence.}
Lookahead layouts improve dynamics, but full-sequence non-causal inference
entangles the current output with unrealized forecasts. In closed-loop rollout,
the lookahead used at step \(t\) may change after the policy receives the
generated observation. Once observations or latent states produced under the old
forecast are reused in later queries, future-conditioned information becomes
part of the committed rollout state, causing state mismatch and accumulated
errors.

We therefore adopt rolling-window diffusion, which we find better suited to
step-wise high-dynamic rendering. Instead of denoising a full future clip at
once, rolling diffusion maintains a FIFO latent window: each roll injects a
new noisy latent, denoises the window, and emits one frame. This rolling form
naturally matches the simulator's step-wise query pattern and avoids committing
an entire future-conditioned sequence. It also allows predicted layouts to guide
near-future dynamics, which improves instance motion compared with causal
full-sequence inference, as shown in Fig.~\ref{fig:observation}-(5). However,
naive rolling still exposes the inherited latent state to predicted layouts,
so future-conditioned information can be carried into the next simulator query.

\textbf{Reset-and-Roll} aligns rolling inference with the closed-loop simulator interface as shown in Fig.~\ref{fig:genframe}.
At each step, the window contains \((L-1)\) history frames and one current-step
latent. After the first denoising update, before injecting any lookahead layout, we cache the current latent and timestep. This cached latent has only
seen the committed history and the current simulator layout
\(\mathcal{L}^{\mathrm{sim}}_t\). We then continue rolling denoising with
predicted layouts \(\hat{\mathcal{L}}_{t+1:t+K}\) to render the current
observation with stronger dynamic cues. Before moving to the next simulator
step, all lookahead-exposed latents are discarded. The next query is formed with
the newly realized layout and the cached latent at the same diffusion scheduler
timestep. Thus, predicted layouts are used for temporary denoising in the current
query, but future-conditioned latent states are not committed across steps.

\subsection{Point Cloud Skeleton}
\label{data}

To address the error accumulation observed in AR generation
(Fig.~\ref{fig:observation}-(5)), we introduce the Point Cloud Skeleton as a
training-inference-consistent scene representation. It accumulates offline LiDAR
observations into static background and track-indexed foreground assets,
which are recomposed by the simulator and projected to camera views at each
rollout step. The projected skeleton provides sparse but state-aligned
current-view evidence, turning each step into conditional rendering rather than
pure history-based extrapolation. Thus, the primary role of the skeleton is to
provide a consistent state representation during rollout, rather
than to reconstruct the scene at the pixel level.

\begin{figure}[!t]
  \includegraphics[width=\textwidth]{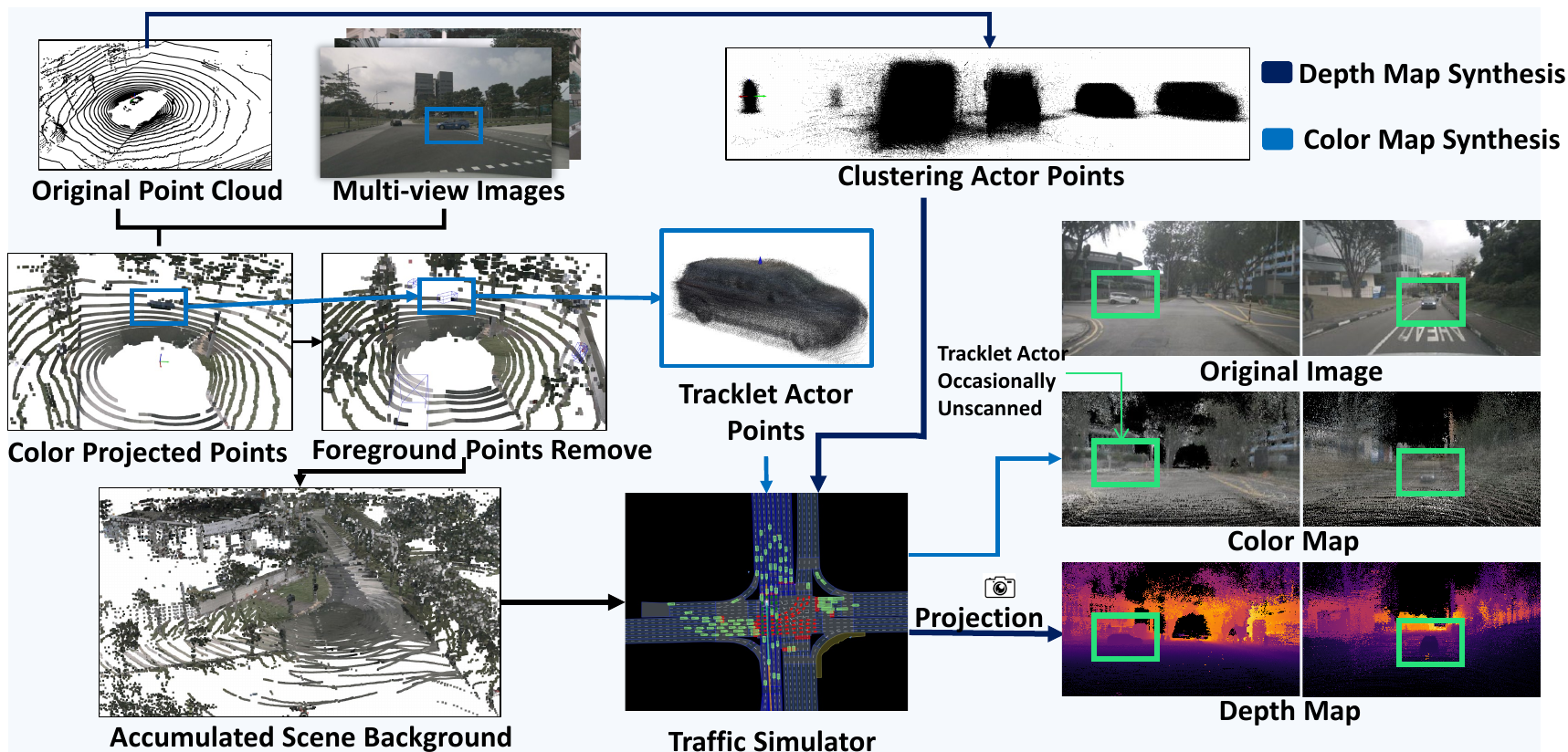}
  \caption{
  \textbf{Construction and Utilization of Point Cloud Skeleton.} We separately accumulate foreground and background skeleton using point cloud and images from recorded logs. During simulation, actor assets are recomposed with the background according to simulator states and projected to camera views. The skeleton provides two conditioning inputs: a color map for appearance anchors and a template-based depth map for view-consistent geometry of moving agents.
  }
  \label{fig:datapipe}
 \vspace{-6mm}
\end{figure}

As illustrated in Fig.~\ref{fig:datapipe}, we colorize each LiDAR frame by
projecting points onto calibrated images, obtaining \(P_t^c\). Using 3D boxes
and track IDs, we split it into background and foreground points:
\begin{equation}
P_t^c = P_t^{\mathrm{bg}} \cup P_t^{\mathrm{fg}}, 
\quad
P_t^{\mathrm{fg}}=\bigcup_{i\in\mathcal{A}_t} P_{t,i}^{\mathrm{fg}},
\quad
P_t^{\mathrm{bg}}\cap P_t^{\mathrm{fg}}=\emptyset ,
\end{equation}
where \(\mathcal{A}_t\) denotes the actors observed at frame \(t\).
We accumulate background points in the global frame, and normalize each actor's
foreground points into canonical coordinates:
\begin{equation}
P_{\mathrm{global}}^{\mathrm{bg}}
=
\bigcup_{t=1}^{N} T_{\mathrm{global}}(t) P_t^{\mathrm{bg}},
\quad
P_{\mathrm{actor}(i)}^{\mathrm{fg}}
=
\bigcup_{t:i\in\mathcal{A}_t} B_{t,i}^{-1} P_{t,i}^{\mathrm{fg}} .
\end{equation}
Here \(T_{\mathrm{global}}(t)\) maps frame-\(t\) points to the global frame, and
\(B_{t,i}\) denotes the 3D box pose of actor \(i\). Thus, \(B_{t,i}^{-1}\)
recenters and aligns foreground points into canonical coordinates, forming a
track-indexed actor asset.
During simulation, the simulator places each actor according to
box pose, merges it with \(P_{\mathrm{global}}^{\mathrm{bg}}\), and projects the
composed skeleton to each camera view as the \emph{color map} condition.

The projected skeleton is used through two complementary branches. The
\emph{color branch} preserves pixel-level appearance from accumulated
colorized points, providing degraded but useful appearance anchors for
autoregressive rendering. However, foreground actors are often only partially
scanned in the logs. Under trajectory drift, their color projections may become
sparse or view-inconsistent, creating a training-inference gap that is amplified
during rollout.
We therefore introduce a \emph{depth branch} to provide more stable geometry.
We cluster training-log 3D boxes by category, and aggregate LiDAR
points within each cluster to build category-level point templates for common traffic participants, such as sedan, SUV, and pedestrian. During simulation,
\textbf{these templates are placed according to simulator-provided actor boxes and
projected as depth maps}. The color map supplies appearance anchors, while the
template-depth map supplies geometry. We concatenate
both maps with the box\&hdmap layout as conditions for the driving world model.

\subsection{nuPlan-SimGen}
\label{sim_geneval}

\begin{figure*}[!t]
  \includegraphics[width=\textwidth]{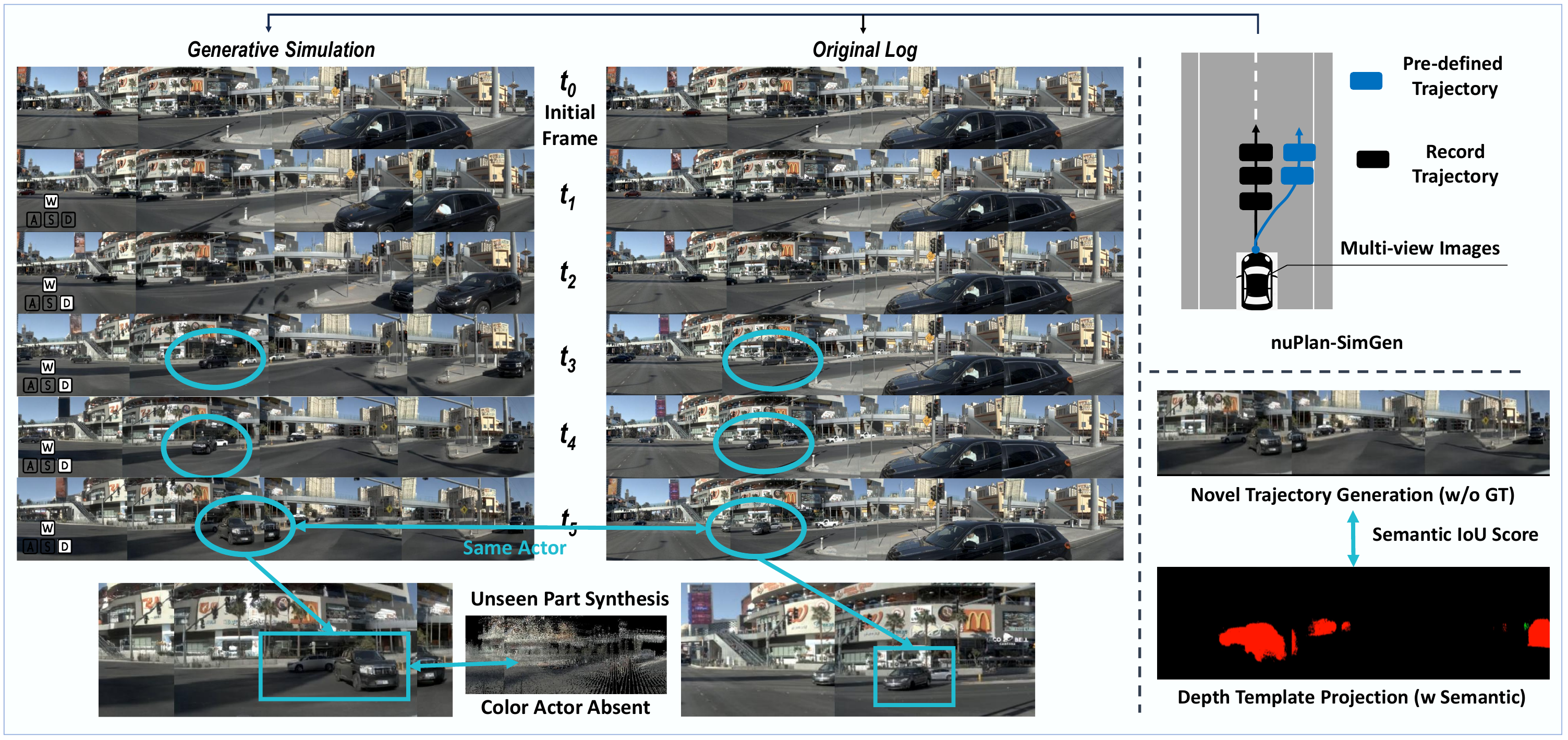}
  \caption{
      \textbf{nuPlan-SimGen.} Starting from GT reference frames, nuPlan-SimGen evaluates generative rendering when a predefined lane-change trajectory takes over the ego motion. Projected point templates provide vehicle pseudo masks for projected-mask IoU under novel trajectories where no image ground truth is available.
  }
  \label{fig:nuplantest}
    \vspace{-6mm}
\end{figure*}

To connect generative models with practical closed-loop simulation, we implement
\textbf{nuPlan-SimGen}, a nuPlan~\cite{nuplan} plugin for closed-loop
generative rendering. The plugin attaches the visual generator to the nuPlan
traffic simulator: after each planner action, the simulator updates ego and
actor states, and the generator renders the corresponding camera observations.
Enabled by our template-based depth map, nuPlan-SimGen also supports a simulator-state alignment proxy under deviated ego trajectories. We project movable-object point templates to obtain vehicle pseudo masks and compare them with vehicle masks predicted by an off-the-shelf segmentor, reporting projected-mask IoU 
(Fig.~\ref{fig:nuplantest}).

As shown in Fig.~\ref{fig:nuplantest}, nuPlan-SimGen evaluates
generative rendering under ego-trajectory drift. During the first \(N\) steps,
the ego vehicle and surrounding actors follow the logged trajectories, where
\(N\) is the number of reference frames required by the generator. Starting from
frame \(N\), a predefined lane-change trajectory takes over the ego motion. This
trajectory gradually adjusts the ego speed toward a target value of 10m/s and applies a
yaw deviation from the recorded path. At each simulator step, the next executable waypoint is committed as the current layout, while the remaining waypoints are treated as future layouts. Surrounding actors can either continue along their
logged tracks or be controlled by nuPlan's IDM policy; for consistency, our
evaluation keeps them on the logged tracks. Although their motions are unchanged,
the deviated ego pose alters relative viewpoints and visibility, which may bring
actors outside the logged camera view into the generated field of view.

\begin{table}[tb!]
    \centering
    \caption{\textbf{Comparison on nuScenes Dataset.}}
    \setlength{\tabcolsep}{16pt}      

    \resizebox{\textwidth}{!}{
    \begin{tabular}{l|c|l|c|cc}
        \toprule
        Method & Backbone & Condition & Resolution
        & $\mathrm{FID}_{(\downarrow)}$ & $\mathrm{FVD}_{(\downarrow)}$ \\
        \midrule
        Panacea+~\cite{panacea+} & SD1.5 & Layout$+$Text & 256$\times$512 & 15.50 & 103.00 \\
        \rowcolor{gray!15} 
        MagicDrive~\cite{magicdrive} & SD1.5 & Layout$+$Text & 224$\times$400 & 16.20 & 217.94 \\
        DriveArena~\cite{drivearena} & SD1.5 & Layout$+$Text$+$Prev & 224$\times$400 & 16.03 & 185.32 \\
        \rowcolor{gray!15} 
        MagicDiT~\cite{magicdrivedit} & Sora-DiT & Layout$+$Text$+$Video & 224$\times$400 & 20.91 & 97.21 \\
        Epona~\cite{epona} & Flux-DiT & Ego Traj$+$Text$+$Video & 512$\times$1024 & 7.50 & 82.80 \\
        \midrule \midrule
        \rowcolor{gray!15} 
        Ours (Image) & SD1.5 & Layout$+$Point$+$Prev & 224$\times$400 & 7.09 & 69.47 \\
        Ours (Video)& SD3.5 & Layout$+$Point$+$Video & 256$\times$448 & \textbf{5.97} & \textbf{58.3} \\
        \bottomrule
    \end{tabular}
    }
    \vspace{-4mm}
    \label{tab:nuscenes_comparison}
\end{table}

\section{Experiments}
\subsection{Datasets and Implementation Details}

We conduct experiments on nuScenes and nuPlan-mini. On nuScenes, we use 700
training and 150 validation scenes, reporting FID, FVD, and UniAD scores~\cite{drivearena,uniad}. On nuPlan-mini, we use all 64 sequences for model training and select 300
customized-trajectory clips (specific in appendix) for nuPlan-SimGen evaluation. This benchmark evaluates trajectory deviation within logged scenes, rather than scene-level generalization to unseen logs or maps. We report FID and Mask2Former vehicle segmentation scores~\cite{masked2former}.

We implement two variants of \emph{Point as Skeleton}: an SDv1.5-based
single-frame generator~\cite{LDM} for comparison with image generators
~\cite{drivearena,you2024bench2drive}, and an SDv3.5-based autoregressive video
generator for comparison with video generators~\cite{epona,dreamforge}. The
SDv1.5 variant is trained for 200 epochs on nuScenes and 150 epochs on nuPlan,
with batch sizes of 16 and 8, respectively. The SDv3.5 variant is trained for
15k iterations with batch size 16 on both datasets. All models use diffusion
forcing~\cite{dfot}. We generate 2 Hz images and 6 Hz videos on nuScenes using
ASAP labels~\cite{ASAP}, and 5 Hz videos on nuPlan. More implementation details
are provided in the appendix.

\subsection{Main Results}
\subsubsection{Quantitative Analysis}

We first evaluate \emph{Point as Skeleton} on nuScenes. In video generation,
the model conditions on 2s history and generates a 4s clip at 6 Hz, yielding
24 autoregressive steps. The image variant is evaluated
for 16 autoregressive steps at 2 Hz.
Table~\ref{tab:nuscenes_comparison} shows the best FID and FVD under this
evaluation protocol, indicating improved visual quality and temporal
consistency. We further
evaluate generated sequences with UniAD. Table~\ref{tab:nus_uniad} shows
stronger downstream detection and segmentation, with competitive planning
performance. These results suggest that \emph{Point as Skeleton} better preserves
perception-relevant information while maintaining competitive planning
performance.

\begin{table}[tb!]
    \centering
    \caption{\textbf{UniAD’s~\cite{uniad} Different Tasks in nuScenes Generation.}}
    \setlength{\tabcolsep}{8pt} 
    \resizebox{\textwidth}{!}{
    \begin{tabular}{c|cc|cccc|ccc}
    \toprule
        \multirow{2}{*}{Source Data} 
        & \multicolumn{2}{c|}{3DOD $\uparrow$} 
        & \multicolumn{4}{c|}{Segmentation $\uparrow$} 
        & \multicolumn{3}{c}{$\text{L2}$ $\downarrow$} \\
        \cline{2-10}
        & mAP & NDS 
        & Lanes & Drivable & Divider & Crossing 
        & 1.0s & 2.0s & 3.0s \\
    \midrule
    ori nuScenes~\cite{nuscenes} & 37.98 & 49.85 & 31.31 & 69.14 & 25.93 & 14.36 & 0.51 & 0.98 & 1.65 \\
    \rowcolor{gray!15} 
    MagicDrive~\cite{magicdrive} & 12.92 & 28.36 & 21.95 & 51.46 & 17.10 & 6.57 & 0.57 & 1.14 & 1.95 \\
    Panacea~\cite{panacea} & 13.72 & 27.73 & 18.23 & 52.37 & 17.21 & 6.32 & 0.58 & 1.14 & 1.97 \\
    \rowcolor{gray!15} 
    DriveArena~\cite{drivearena} & 16.06 & 30.03 & 26.14 & 59.37 & 20.79 & 8.92 & 0.56 & 1.10 & 1.89 \\
    \midrule \midrule
    Ours (Image) & 15.82 & 32.87 & 26.37 & 57.66 & 21.59 & 9.18 & \textbf{0.53} & \textbf{1.06} & 1.84 \\
    \rowcolor{gray!15} 
    Ours (Video) & \textbf{25.59} & \textbf{40.93} & \textbf{28.76} & \textbf{63.76} & \textbf{23.69} & \textbf{12.04} & 0.56 & 1.10 & \textbf{1.83} \\
    \bottomrule
    \end{tabular}}
    \vspace{-4mm}
    \label{tab:nus_uniad}
\end{table}

\begin{figure*}[!t]
\centering
\includegraphics[width=\textwidth]{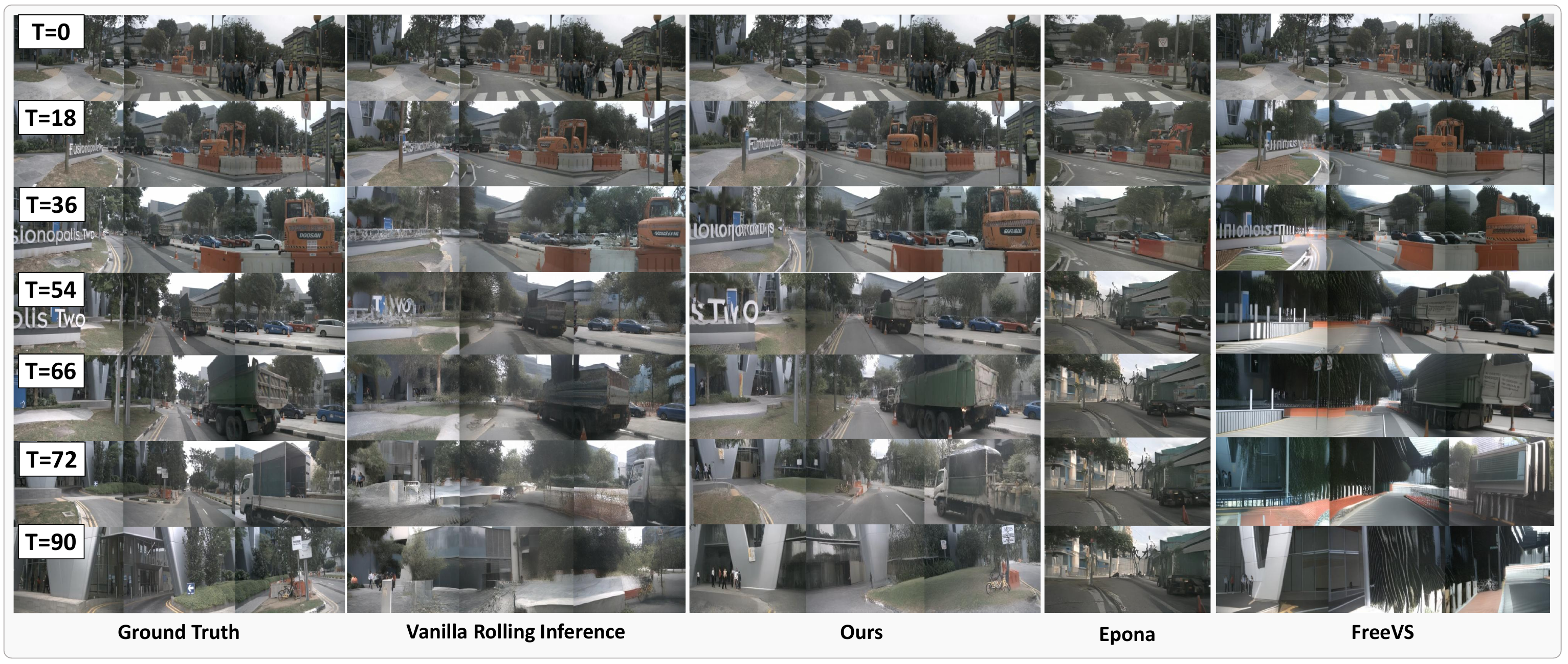}
    \caption{\textbf{Visualization of Long Horizon Rollout.} FreeVS~\cite{freevs} is reproduced in a non-causal full-sequence setting. Our method is better suited to simulation scenarios than the compared methods.}
    \label{fig:qualitative}
    \vspace{-6mm}
\end{figure*}

\begin{table}[H]
\centering

\begin{minipage}[t]{0.48\columnwidth}
\vspace{0pt}
\vspace{-3mm}
    \centering
    \footnotesize
    \setlength{\tabcolsep}{5pt}
    \captionof{table}{\textbf{Results on the Interactive Generation.}
    *FreeVS is reproduced in an autoregressive setting using the same resolution and backbone.}
    \label{tab:dynamic}
    \resizebox{\linewidth}{!}{%
    \begin{tabular}{l|l|c|c}
    \toprule
    Method & Design Mode & FID$\downarrow$ & IoU(\%)$\uparrow$ \\
    \midrule
    MagicDrive~\cite{magicdrive} & Image & 47.15 & 42.47 \\
    \rowcolor{gray!15}
    DriveArena~\cite{drivearena} & Autoregressive Image & 45.38 & 46.34 \\
    FreeVS*~\cite{freevs} & Video & 27.37 & 54.72 \\
    \midrule
    \rowcolor{gray!15}
    Ours (Image) & Autoregressive Image & 27.96 & 58.57 \\
    Ours (Video) & Autoregressive Video & \textbf{18.37} & \textbf{59.92} \\
    \bottomrule
    \end{tabular}}
\end{minipage}
\hfill
\begin{minipage}[t]{0.50\columnwidth}
\vspace{0pt}
    \textbf{nuPlan-SimGen.}
    We evaluate closed-loop generation in Table~\ref{tab:dynamic}, and report FID together with point-label
    IoU. Since the ego trajectory deviates from the log, no image ground truth is
    available and FID is only an auxiliary metric. In contrast, projected-mask IoU
    measures whether generated vehicles align with simulator-provided geometry. Our
    method achieves the best IoU, suggesting better closed-loop alignment.
\end{minipage}

\vspace{-4mm}
\end{table}

\subsubsection{Qualitative Analysis.}
As shown in Fig.~\ref{fig:nuplantest}, \emph{Point as Skeleton} can synthesize long-horizon interactive multi-view camera sequences. It offers a simple way to handle unobserved actor parts during temporal extrapolation and allows new actions to be inserted at each rollout step.
We further evaluate long-horizon generation on nuScenes in Fig.~\ref{fig:qualitative}. We roll out 100 consecutive frames at 6 Hz, covering nearly the full nuScenes validation sequence length of 17 seconds. Compared with prior methods~\cite{epona,drivearena,freevs}, our design is better suited to simulation-oriented rollouts. It also remains more stable than autoregressive generation without point-skeleton conditioning, especially when the ego vehicle moves beyond initially observed regions (T40--70) or encounters rare street layouts (T80--90).

\begin{figure*}[!t]
\centering
\includegraphics[width=\textwidth]{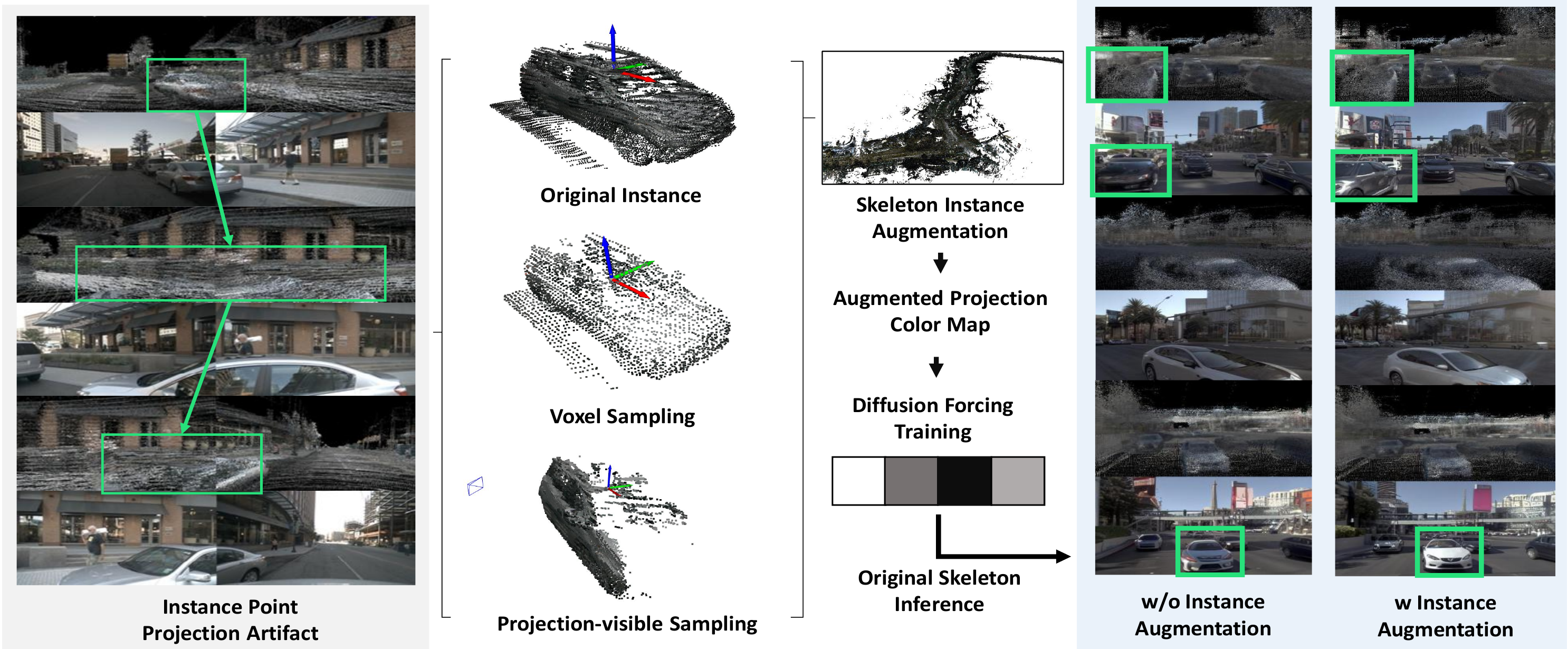}
    \caption{\textbf{3D Instance Augmentation Pipeline.}}
    \label{fig:instanceaug}
        \vspace{-3mm}
\end{figure*}

\begin{figure*}[!t]
\centering
\begin{minipage}[t]{0.47\textwidth}
    \vspace{0pt}
    \centering
    \includegraphics[width=\linewidth]{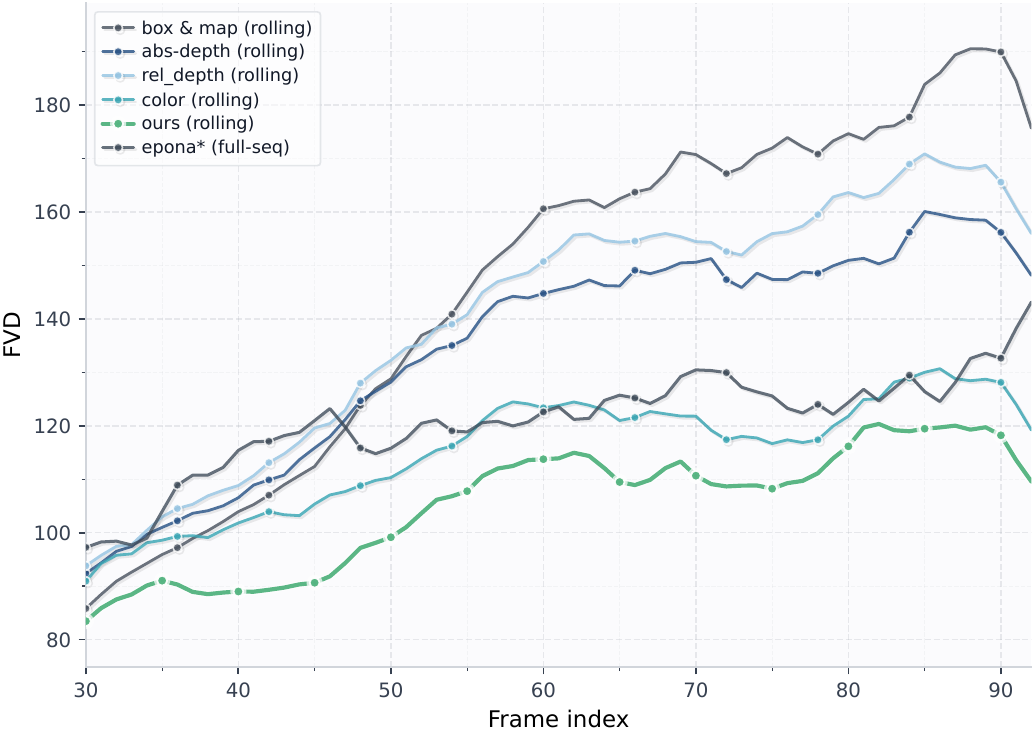}
    \captionof{figure}{
    \textbf{Framewise-AR Comparison.}
    We compute 15-frame window FVD with stride 1. Since Epona* outputs only the front view, we replicate it to 6 views for evaluation.
    }
    \label{fig:curve}
\end{minipage}%
\hfill%
\begin{minipage}[t]{0.49\textwidth}
    \vspace{0pt}
    \centering

    \setlength{\tabcolsep}{4pt}
    \setlength{\aboverulesep}{0.35ex}
    \setlength{\belowrulesep}{0.35ex}
    \setlength{\cmidrulesep}{0.10ex}

    \captionof{table}{\textbf{Condition ablation on nuPlan.}
    LO/P/$\tilde{P}$: layout/previous/noisy latent;
    C/D: color/depth condition; Aug(C): color augmentation.}
    \label{tab:ablation_exp}
    \vspace{-1mm}

    \begin{tabular}{@{\hspace{2.5mm}}p{0.50\linewidth}
                    >{\centering\arraybackslash}p{0.17\linewidth}
                    >{\centering\arraybackslash}p{0.17\linewidth}@{\hspace{2.5mm}}}
        \toprule
        Cond. & FID$\downarrow$ & IoU$\uparrow$ \\
        \midrule
    
        \multicolumn{3}{@{}l}{\textit{Color conditioning}} \\[-0.5mm]
        \cmidrule(lr){1-3}
        LO+P   & 44.88 & 46.89 \\
        \rowcolor{gray!15}
        LO+P+C & \textbf{38.27} & \textbf{50.91} \\
        \midrule
    
        \multicolumn{3}{@{}l}{\textit{Template depth}} \\[-0.5mm]
        \cmidrule(lr){1-3}
        LO+$\tilde{P}$+C   & \textbf{27.82} & 54.52 \\
        \rowcolor{gray!15}
        LO+$\tilde{P}$+C+D & 28.45 & \textbf{55.67} \\
        \midrule
    
        \multicolumn{3}{@{}l}{\textit{Color aug. with depth}} \\[-0.5mm]
        \cmidrule(lr){1-3}
        LO+$\tilde{P}$+C+Aug(C)   & 28.32 & 56.19 \\
        \rowcolor{gray!15}
        LO+$\tilde{P}$+C+D+Aug(C) & \textbf{27.85} & \textbf{58.57} \\
        \bottomrule
    \end{tabular}
    \end{minipage}

\vspace{-6mm}
\end{figure*}

\subsection{Ablation Study}

\textbf{Ablation of Conditions.}
We ablate Point Skeleton conditioning on nuScenes and nuPlan. On nuScenes, we compare box \& map, relative depth with denser near-range bins, raw metric depth, color, and metric depth+color for 100-frame AR generation, as shown in Fig.~\ref{fig:curve}. Metric depth provides stronger guidance than relative depth, while color further stabilizes synthesis quality; combining color and metric depth achieves the best result. The FVD drop around frame 90 is caused by tail-flush generation, where the rolling window stops and stays fixed until all frames are fully denoised.
On nuPlan, we study short-clip arbitrary-trajectory generation in Tab.~\ref{tab:ablation_exp} using image generation model. Here, $\tilde{P}$ denotes a previous latent perturbed by 50--200 diffusion steps during training. Diffusion forcing improves robustness, and projected color conditioning further improves both FID and IoU. Directly appending metric depth to color brings only marginal gains, as the depth signal can be masked by dominant appearance and history cues. However, with foreground-instance augmentation, color+depth achieves the best overall result, indicating that depth provides complementary geometric guidance when color cues degrade during autoregressive inference.

\vspace{-2mm}
\paragraph{Instance augmentation.}
\label{abl-instance}
As shown in Fig.~\ref{fig:instanceaug}, foreground vehicles are more prone to projection artifacts than static backgrounds when the ego pose and viewing direction change. This introduces a training-validation gap: during training, point projections are usually aligned with logged camera views, whereas arbitrary simulation trajectories can expose unseen views of nearby agents. To mitigate this mismatch, we augment foreground instances during training by corrupting 80\% of instance assets with either voxelization or simulated camera projections, and mixing them with clean assets. Since the sparse point clouds in nuScenes naturally reduce this gap, we evaluate this augmentation mainly on nuPlan. The last block of Tab.~\ref{tab:ablation_exp} and Fig.~\ref{fig:instanceaug} show that instance augmentation improves both quantitative performance and visual consistency.
\vspace{-2mm}

\section{Conclusion}
\vspace{-2mm}
We present \textit{Point as Skeleton}, a point-cloud-conditioned generative
simulator for closed-loop-compatible autonomous driving rendering. Reset-and-Roll
separates committed simulator states from transient forecasts, allowing rolling
denoising to use lookahead guidance without carrying future-conditioned latents
across simulation steps. The point-cloud skeleton further decomposes logged
LiDAR observations into static background and track-indexed foreground assets,
which are projected as color and template-depth conditions to anchor
autoregressive rollout. Experiments on nuScenes and nuPlan show improved visual
realism and geometric alignment, demonstrating the potential
of integrating autoregressive generative models into closed-loop driving rollout.

\bibliographystyle{unsrtnat}
\bibliography{main}

\appendix

\newpage

\section{Visualization}
\begin{figure}[t!]
\includegraphics[width=\textwidth]{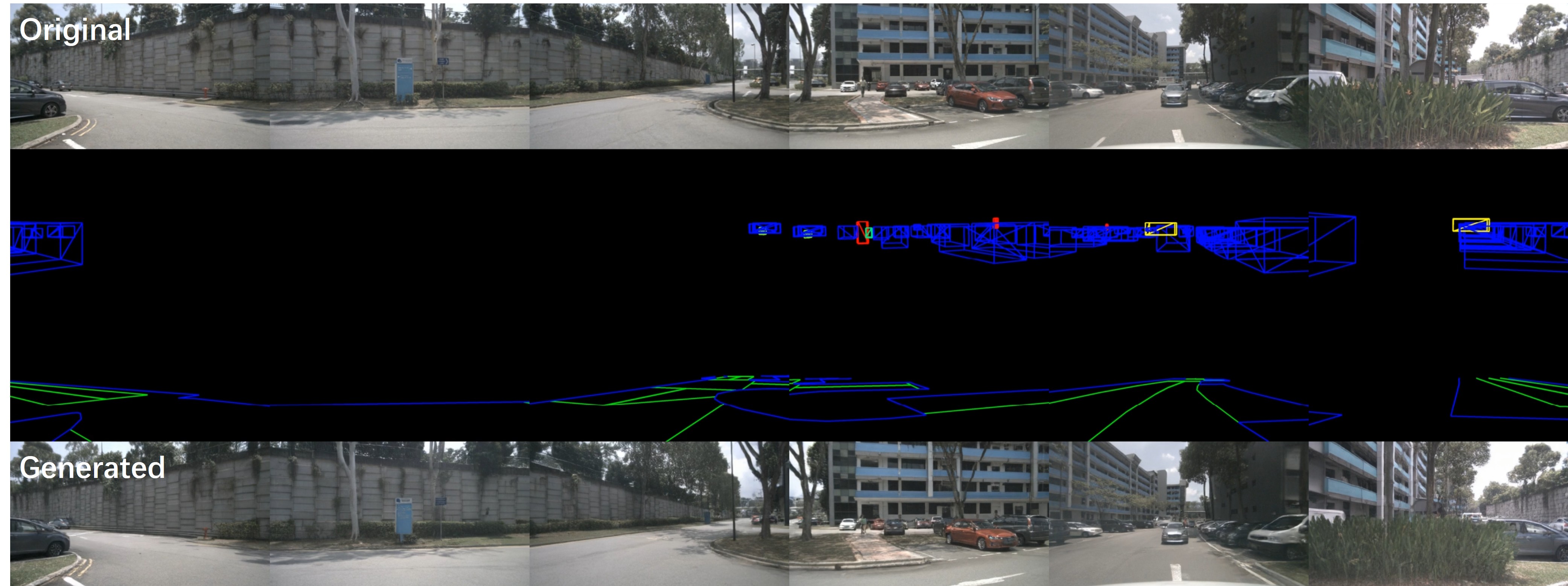}
  \caption{
  \textbf{Long Video Generation.} A 17-second generation produces a 100-frame video, including 89 autoregressive steps after excluding the 11 initial frames. This long video demonstrates that the point-as-skeleton representation supports a generation length sufficient to complete a full sequence of actions.
  }
  \vspace{-2mm}
  \label{fig:long}
\end{figure}

\begin{figure}[t!]
\includegraphics[width=\textwidth]{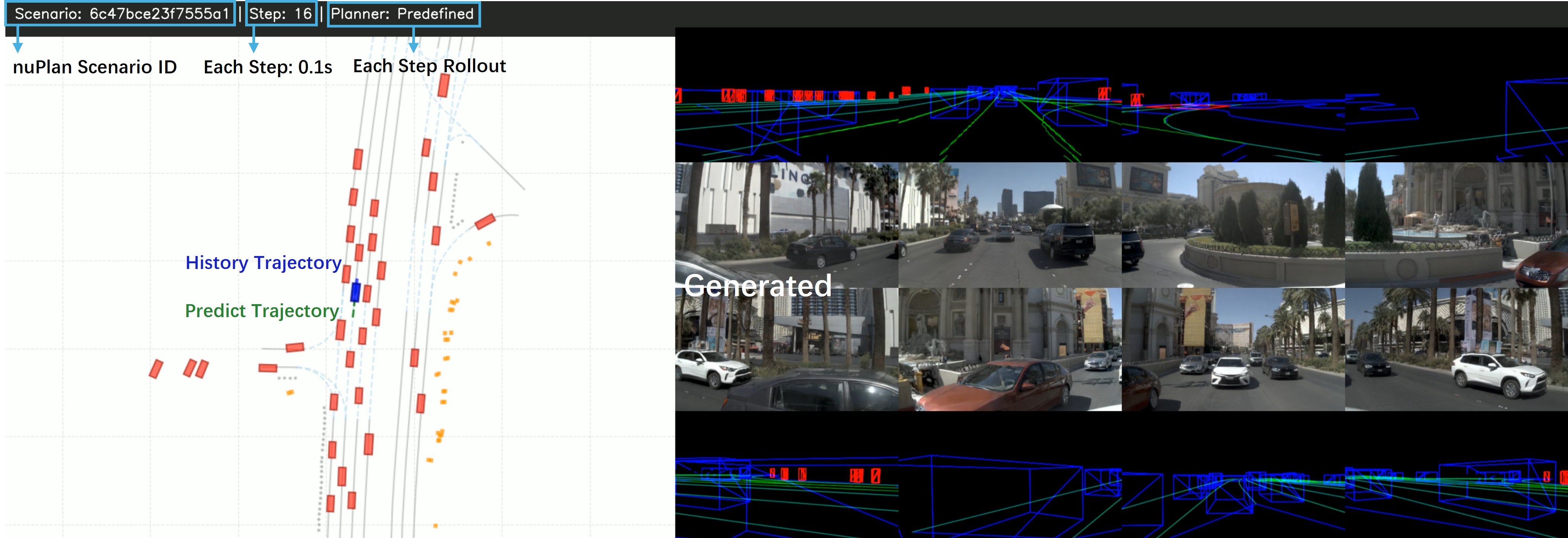}
  \caption{
  \textbf{nuPlan Plugin.} Starting from step 16, we perform a rollout every 0.2 seconds for trajectory control and generate one frame at each rollout. This video demonstrates that our method can be seamlessly integrated with modern driving planning systems.
  }
  \vspace{-2mm}
  \label{fig:nuplan}
\end{figure}

All our videos are generated in a frame-by-frame autoregressive manner. Before generating each frame, the motion trajectory can be specified by the user, which enables \textbf{interactive generation}. The nuScenes videos are generated at 6 Hz, while the nuPlan videos are generated at 5 Hz. The details explanation of the videos are shown in the Fig.~\ref{fig:long} and Fig.~\ref{fig:nuplan}. Note that the stored video loses some degree of visual clarity.

\section{Discussion}

\noindent
\begin{minipage}[t]{0.3\linewidth}
    \vspace{0pt}
    \centering
    \includegraphics[width=\linewidth,page=1]{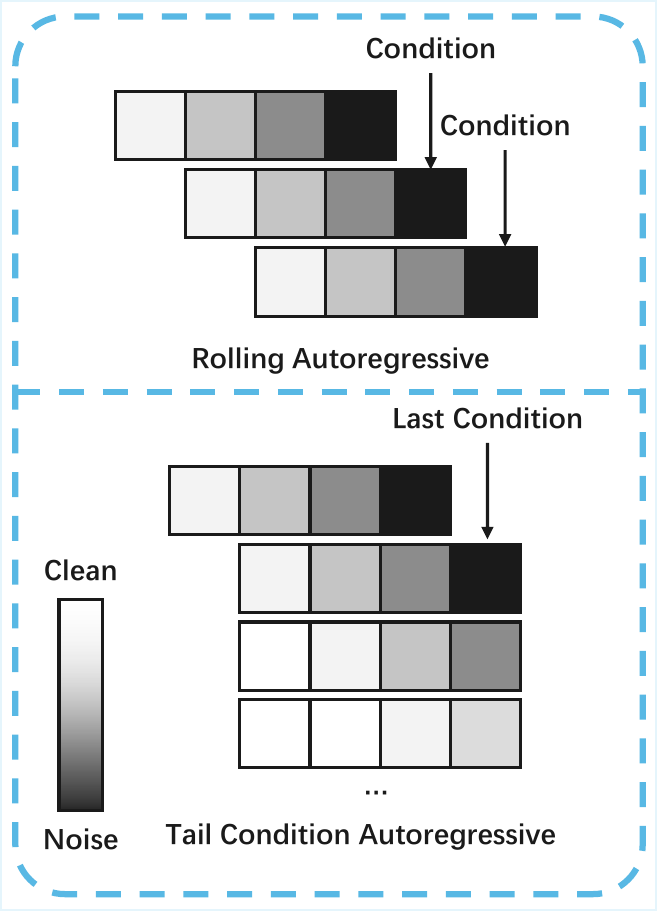}
    \captionof{figure}{\textbf{Tail Generation of Fixed-Length Condition.}}
    \label{fig:disscus}
\end{minipage}\hfill
\begin{minipage}[t]{0.66\linewidth}
    \vspace{0pt}
    \small
    In Fig.~9, all rolling-based methods exhibit a decreasing trend in tail FVD, which is inconsistent with the common understanding of autoregressive error accumulation. We find that this phenomenon is specific to fixed-length, fixed-trajectory video generation on nuScenes, and does not arise in the nuPlan simulation. As shown in the Fig.~\ref{fig:disscus}, when the rolling window reaches the end of the video (17.5\,s), the nuScenes dataset cannot provide additional trajectory conditions for most sequences. We therefore keep the window fixed in place to complete the remaining generation. As a result, this stage follows a generation pattern different from the normal rolling process, which rollout future trajectory in each step.

    Although we have some intuition about this phenomenon, the clean frame sink design~\cite{diffusionforcing} remains an open question. While it may improve short-term video quality, its impact on motion fidelity and error accumulation in autonomous driving simulation deserve systemically exploration.
\end{minipage}


\section{Derivations}

\paragraph{\textbf{Causal full-sequence tends to static instances.}}
Without future layout cues, the next-frame conditional is a marginal over future layouts:
\begin{equation}
p(x_{t+1}\mid h_t)=\int p(x_{t+1}\mid h_t,f_t)\,p(f_t\mid h_t)\,df_t .
\label{eq:app_causal_marginal}
\end{equation}

Let $z_s$ denote the diffusion latent at noise level $s$. Under the standard forward perturbation, we have
\begin{subequations}
\label{eq:app_z_mixture}
\begin{equation}
z_s=\sqrt{\bar\alpha_s}\,x_{t+1}+\sqrt{1-\bar\alpha_s}\,\epsilon,
\qquad \epsilon\sim\mathcal{N}(0,I) .
\label{eq:app_z_def}
\end{equation}
\begin{equation}
p(z_s\mid h_t)=\int p(z_s\mid h_t,f_t)\,p(f_t\mid h_t)\,df_t ,
\label{eq:app_z_marg}
\end{equation}
\end{subequations}
where $p(z_s\mid h_t,f_t)$ is the distribution induced by forward diffusion from $p(x_{t+1}\mid h_t,f_t)$.

In conditional diffusion score matching, the optimal denoising direction is the conditional score
$s^{*}(z_s;h_t):=\nabla_{z_s}\log p(z_s\mid h_t)$, which for the above mixture satisfies
\begin{equation}
\nabla_{z_s}\log p(z_s\mid h_t)
=
\mathbb{E}_{f_t\sim p(f_t\mid z_s,h_t)}
\big[\nabla_{z_s}\log p(z_s\mid h_t,f_t)\big] .
\label{eq:app_score_average}
\end{equation}
Eq.~\eqref{eq:app_score_average} shows that, without future layout cues, the optimal denoising direction becomes a
posterior-weighted \emph{average} over future-conditioned modes. When $p(f_t\mid z_s,h_t)$ is multi-modal (e.g., at
low-speed or stop-and-go where multiple motion hypotheses are plausible), mode-wise denoising directions may partially
cancel, yielding conservative updates that favor persistence of current instances and leading to static agents in
rollouts. Conversely, when future motion is near-deterministic, $p(f_t\mid z_s,h_t)$ concentrates and the averaging
effect diminishes.

\section{Implementation Details}

\textbf{Autoregressive Generator.} For the nuPlan-mini dataset, we use a fixed prompt template with a variable weather description: “{weather}. A driving scene in {city}”, where the weather term is instantiated for each sample. For nuScenes, we adopt the text annotations from OpenDWM~\cite{unimlvg}. The model is trained using the AdamW optimizer with a learning rate of 6e-5. For the video setting, we use an autoregressive window of 2 seconds, corresponding to 12 frames at 6 Hz, with 11 initial reference frames. For the image setting, only a 2-frame window is used. Note that the FVD for long-video generation is computed from saved MP4 files, which may introduce a slight loss in visual quality due to compression.

\textbf{Point Cloud Skeleton.} We apply voxel downsampling with a voxel size of 0.01m to the stacked city background point clouds, while the tracklet actors remain unchanged. For depth-map synthesis, we mainly use vehicle-size prototype templates as foreground depth hints, while during nuPlan-SimGenEval we use tracklet actor first if the points number is higher than 20000. During the simulation, we only consider point clouds within a 100m radius of the ego vehicle. We implement CUDA-based rasterization to accelerate point projection in
nuPlan-SimGenEval.

\textbf{Interactive Divergence Track Construction.} For nuPlan-mini, we uniformly sample data instances such that each video sequence of approximately 400\,s can be divided into about 30 clips. We first pretrain the layout-based autoregressive model on the text-annotated nuPlan-mini set using all available splits, and then finetune it on 12 subsets with point-skeleton. Finally, we select 10 subsets for customized-trajectory evaluation. In the trajectory-customization test, we generate camera data corresponding to a right-lane-change maneuver of the ego vehicle. The point mask for segment GT applies Gaussian blur with a kernel size of 3 to make the projected point cloud denser in the image plane.

For the IoU evaluation, we assign a category label to each decoupled actor point when constructing the point-cloud skeleton. During rasterized point-to-image projection, these point-wise labels serve as point-template pseudo labels for evaluating alignment with simulator-provided actor geometry. For evaluation, we use an official Mask2Former checkpoint trained on the Cityscapes dataset and report segmentation scores on the \emph{Vehicle} category. The public Mask2Former checkpoint is fixed across all methods, so the metric is used as a consistent geometry-alignment proxy rather than image-grounded semantic ground truth.
The evaluation subsets used for customized-trajectory nuPlan-SimGenEval are:
\begin{itemize}
    \item \texttt{2021.05.12.22.28.35\_veh-35\_00620\_01164}
    \item \texttt{2021.05.12.22.00.38\_veh-35\_01008\_01518}
    \item \texttt{2021.05.12.23.36.44\_veh-35\_02035\_02387}
    \item \texttt{2021.05.12.23.36.44\_veh-35\_01133\_01535}
    \item \texttt{2021.05.12.23.36.44\_veh-35\_00152\_00504}
    \item \texttt{2021.05.25.14.16.10\_veh-35\_01690\_02183}
    \item \texttt{2021.06.09.12.39.51\_veh-26\_01943\_02303}
    \item \texttt{2021.06.09.14.58.55\_veh-35\_01894\_02311}
    \item \texttt{2021.06.09.17.23.18\_veh-38\_00773\_01140}
    \item \texttt{2021.06.14.16.32.09\_veh-35\_05038\_05402}
\end{itemize}

\end{document}